\documentclass[10pt,twocolumn,letterpaper]{article}
\usepackage[T1]{fontenc}
\usepackage{pbsi}
\usepackage{cvpr}
\usepackage{graphicx}
\usepackage{subcaption}
\usepackage[accsupp]{axessibility}
\usepackage{booktabs}
\usepackage{amsmath}
\usepackage{multirow} 
\graphicspath{{figs/}{figs/highinfo/}}
\usepackage{textcomp} 
\usepackage[table]{xcolor} 
\usepackage{mathrsfs} 
\definecolor{hdrR}{HTML}{AD1457}   
\definecolor{hdrC}{HTML}{1565C0}   
\definecolor{hdrI}{HTML}{00695C}   
\definecolor{hdrT}{HTML}{E65100}   
\definecolor{bestG}{HTML}{C8E6C9}  
\definecolor{oursR}{HTML}{E3F2FD}  
\definecolor{rowA}{HTML}{F5F5F5}   
\definecolor{pinklink}{RGB}{255,105,180} 
\newcommand{\best}[1]{\cellcolor{bestG}\textbf{#1}}

\definecolor{cvprblue}{rgb}{0.21,0.49,0.74}
\usepackage[pagebackref,breaklinks,colorlinks,allcolors=cvprblue]{hyperref}

\newcommand{\OurModel}{\emph{Sketch2Colab}\xspace}
\newcommand{\ourmodel}{\emph{Sketch2Colab}\xspace}

\title{\texttt{Sketch2Colab}: Sketch-Conditioned Multi-Human Animation\\ via Controllable Flow Distillation\\ [-1ex]}

\author{
    Divyanshu Daiya and 
    Aniket Bera\\
    {\small \textsf{IDEAS Lab, Department of Computer Science, Purdue University}}\\
    {\tt\small \href{mailto:divyanshu@purdue.edu}{divyanshu@purdue.edu}, \href{mailto:aniketbera@purdue.edu}{aniketbera@purdue.edu}}
}
\author{
    Divyanshu Daiya \quad 
    Aniket Bera\\
    {\small \textsf{IDEAS Lab, Department of Computer Science, Purdue University}}\\
    {\tt\small {divyanshu@purdue.edu}, {aniketbera@purdue.edu}}
}

\begin{document}
\twocolumn[{%
\maketitle
\begin{center}
  \centering
  \captionsetup{type=figure}
  \vspace{-6mm}
  \includegraphics[width=\linewidth,page=1]{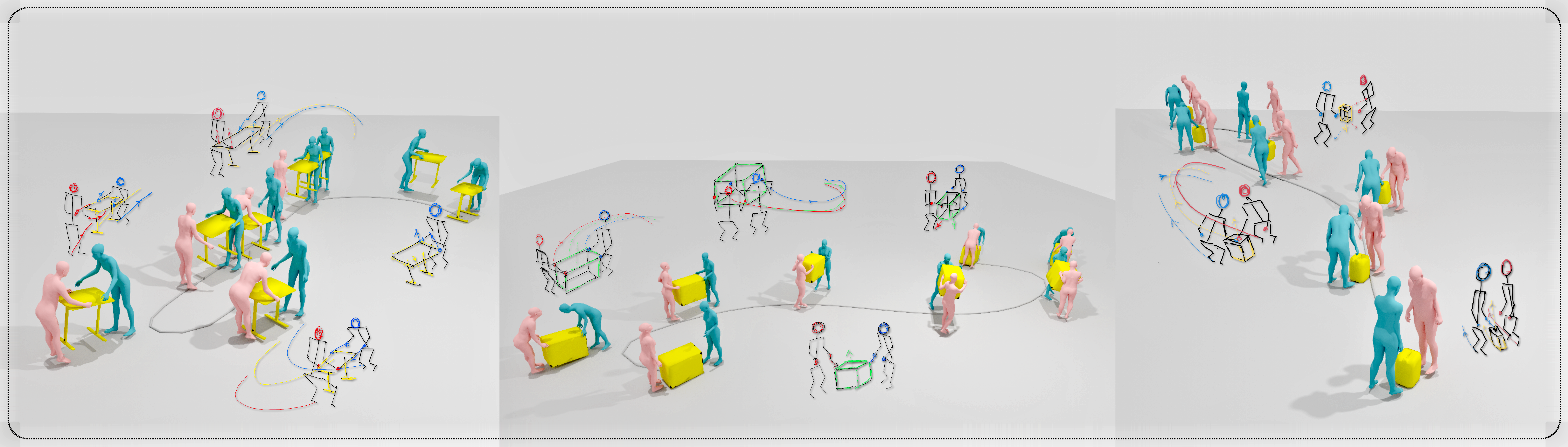}
  \vspace{-6mm}
  \caption{\textbf{Sketch-conditioned human–object–human (HOH) demonstrations with \OurModel.}
Left→right: \textbf{(a)} Two people move a table along a sketched path, then one lets go while the other completes the placement. \textbf{(b)} Two people carry a large box along a prescribed path, adjust its height, and place it down. \textbf{(c)} A specified hand grasps a canister and follows a complex path. All examples use only sparse storyboard keyframes without text. The generated motions follow the intended timing, trajectories, and object contacts.}
  \label{fig:demo}
\end{center} 
}] 
\begin{abstract}
We present \ourmodel, which turns storyboard-style 2D sketches into coherent, object-aware 3D multi-human motion with fine-grained control over agents, joints, timing, and contacts. Diffusion-based motion generators offer strong realism but often rely on costly guidance for multi-entity control and degrade under strong conditioning. \ourmodel instead learns a sketch-conditioned diffusion prior and distills it into a rectified-flow student in latent space for fast, stable sampling. To make motion follow storyboards closely, we guide the student with differentiable objectives that enforce keyframes, paths, contacts, and physical consistency. Collaborative motion naturally involves discrete changes in interaction, such as converging, forming contact, cooperative transport, or disengaging, and a continuous flow alone struggles to sequence these shifts cleanly. We address this with a lightweight continuous-time Markov chain (CTMC) planner that tracks the active interaction regime and modulates the flow to produce clearer, synchronized coordination in human-object-human motion. Experiments on CORE4D and InterHuman show that \ourmodel outperforms baselines in constraint adherence and perceptual quality while sampling substantially faster than diffusion-only alternatives.
\end{abstract}
\vspace{-6mm}

\section{Introduction}
\label{sec:intro}
\vspace{-1mm}
Creating believable multi-entity motion remains labor intensive in game development, film, and VR or AR. Complex scenes, where multiple humans coordinate while interacting with objects, are still largely crafted by hand, requiring long manual cleanup for jitter, collisions, and timing, or motion-capture pipelines that are difficult to scale across tasks and layouts \cite{guo2022generating,tevet2023human}. In recent years, generative motion modeling has progressed rapidly: text-conditioned generation \cite{tevet2023human,zhang2022motiondiffuse,tevet2022motionclip,jiang2023motiongpt,chen2024motionclr}, trajectory or pose-conditioned control and editing \cite{xie2023omnicontrol,karunratanakul2023gmd,rempe2023trace,wan2023tlcontrol,cohan2024flexible,athanasiou2024motionfix}, and style transfer \cite{Song_2024_CVPR,zhong2025smoodi,aberman2020unpaired} have all seen strong results in single-human and, more recently, multi-human or human–scene settings \cite{collage,liu2023interactive,wang2023intercontrol,ghosh2023remos,huang2023diffusion,jiang2024scaling,cong2024laserhuman,gupta2025mdd,gupta2025unified}. Yet comparatively few methods address multi-entity scenarios where several agents coordinate with shared objects under rich task constraints. \textsc{COLLAGE} \cite{collage} represents an early effort in this direction, employing LLM-based planning and a latent diffusion generator to follow complex instructions for human–object–human (HOH) interactions.\\
While text is convenient for planning, it is often an imprecise control channel. Timing, phasing, and spatial layout are more naturally conveyed by visual cues such as keyframes, coarse joint trajectories, and object placement cues. Sketch-based interfaces therefore provide a compelling alternative. \textsc{Sketch2Anim} \cite{s2a} demonstrates that storyboard keyposes and joint or endpoint trajectories can effectively constrain a diffusion model to produce a single-human motion consistent with the artist’s intent. However, \cite{s2a} focuses on one character and does not model inter-agent coordination, hand–object contacts, or collision-free co-manipulation. In this work we ask: \emph{how can we extend sketch-driven control to multi-human, object-centric interactions with coupled constraints?}

Diffusion models are the de facto backbone for motion generation due to their sample quality and data efficiency \cite{ho2020denoising,song2020denoising,dhariwal2021diffusion,tevet2023human,chen2023executing}. Yet achieving \emph{precise} constraint following typically requires task-specific guidance or specialized control modules at inference, which can cause sampling slowdowns, lagging responses to hard constraints, or mode bias toward easy-to-satisfy solutions \cite{chung2022diffusion,xie2023omnicontrol,cohan2024flexible,zhong2025smoodi,gupta2025unified}. Flow-based generators, particularly rectified or conditional flow matching, offer more direct transport under strong conditioning and faster sampling \cite{liu2022flow,lipman2022flow,esser2024scaling,motionlab}. MotionLab \cite{motionlab} shows that rectified flows can unify motion generation and editing with strong conditional alignment.

A straightforward idea, then, is to translate sketch-conditioned diffusion methods, such as \textsc{COLLAGE}-style latent diffusion \cite{collage} or \textsc{Sketch2Anim} \cite{s2a}, into a multi-entity \emph{flow} generator that ingests multi-track keyframes and joint trajectories. However, in our experiments, directly scaling diffusion baselines to multi-entity sketches produced suboptimal adherence, including keyframe misalignments, inter-agent phase drift, and noticeable jitter or lag relative to the target trajectories. Training a competitive flow model \emph{from scratch} for interactive HOH scenarios is also challenging and compute-intensive, as the model must learn a transport field that simultaneously handles multi-agent timing, contacts, and mutual constraints \cite{liu2022flow,lipman2022flow,albergo2023building}. To address this, we adopt a diffusion-to-flow transfer strategy, initializing a rectified-flow student from a strong sketch-conditioned diffusion teacher, in the spirit of recent work on fast rectified-flow sampling and controlled generation \cite{ben2024d,esser2024scaling,motionlab}.

Distillation alone is not sufficient: if the diffusion teacher only weakly satisfies constraints, the student can inherit failure modes. We augment the student with \emph{energy-based} guidance that explicitly shapes the data manifold toward constraint satisfaction. Inspired by recent energy-matching perspectives \cite{energy_work} and classical EBM conditioning \cite{du2019implicit,song2021train}, we define differentiable energies for keyframe pose fidelity, joint or endpoint trajectory tracking, inter-agent or object distances, contact consistency, and collision penalties. During training and sampling, the rectified flow follows transport directions jointly informed by the distilled velocity field and energy gradients, improving adherence without requiring slow posterior guidance loops \cite{chung2022diffusion,ben2024d}.

Finally, multi-entity interaction involves discrete latent events, such as contact on or off, grasp state, and object handoff, whose schedules are difficult to optimize with a purely continuous field. We introduce a lightweight \emph{CTMC-based} planner over interaction states, trained with a Kolmogorov-residual objective inspired by recent generator-learning methods for Markov processes \cite{CTMC}. The CTMC proposes contact and handoff schedules consistent with the sketch, while the continuous rectified flow handles pose-level transport. We couple the two through occupancy-weighted phase mixing and contact-energy gating, which reduces mode switches, temporal lag, and contact flicker.
\vspace{-\baselineskip}
\paragraph{Contributions.} We present a sketch-conditioned, multi-entity motion generator \textbf{\ourmodel} for collaborative HOH scenarios that:
\begin{itemize}
    \item Distills a rectified-flow student from a strong sketch-conditioned diffusion teacher to obtain fast, stable transport under complex conditioning.
    \item Introduces \emph{energy-guided rectified flows} for multi-entity motion: differentiable energies enforce keyframe or trajectory constraints, contacts, and collision avoidance, improving alignment without posterior guidance \cite{energy_work,du2019implicit,chung2022diffusion}.
    \item Integrates a \emph{CTMC-based} phase planner that models approach, contact, and handoff timing, and couples these phases to the continuous flow through phase mixing and contact gating for more reliable multi-agent coordination.
    \item Outperforms the evaluated baselines in constraint adherence and perceptual quality on multi-human, object-centric tasks while sampling substantially faster than diffusion-only alternatives.
\end{itemize} 
\vspace{-2mm}
\section{Related Work}
\label{sec:related}
\vspace{-1mm}
\noindent\textbf{Text-, Trajectory-, and Style-Conditioned Motion.}
Action-conditioned and past-motion models laid early foundations \cite{guo2020action2motion,petrovich2021action,lee2023multiact,athanasiou2022teach,yuan2020dlow,mao2021generating,barquero2022belfusion,chen2023humanmac}. Modern diffusion and GPT-style models enable text-to-motion \cite{tevet2023human,zhang2022motiondiffuse,tevet2022motionclip,jiang2023motiongpt,zhang2023generating,chen2024motionclr}, trajectory control or editing \cite{xie2023omnicontrol,karunratanakul2023gmd,rempe2023trace,wan2023tlcontrol,cohan2024flexible,athanasiou2024motionfix}, and style transfer \cite{Song_2024_CVPR,zhong2025smoodi,aberman2020unpaired}, and extend to multi-person or human–scene settings \cite{liu2023interactive,wang2023intercontrol,ghosh2023remos,huang2023diffusion,li2023object,li2025controllable,jiang2024scaling,cong2024laserhuman}. Our work targets collaborative HOH scenes where multiple agents and shared objects must satisfy coupled constraints.

\noindent\textbf{Collaborative HOH and LLM-Guided Planning.}
\textsc{COLLAGE} \cite{collage} combines LLM planning \cite{chatgpt,touvron2023llama,kojima2022large,huang2022language} with latent diffusion for multi-agent collaboration on datasets such as CORE-4D and InterHuman \cite{zhang2024core4d,liang2023intergen}. While text conveys task structure, it under-specifies timing and spatial alignment, motivating our sketch-based control.

\noindent\textbf{Sketch- and Keyframe-Driven Animation.}
Storyboard sketches encode keyposes and trajectories \cite{lasseter1998principles,williams2012animator}. Prior sketch-to-pose and path-based interfaces \cite{lin2010sketching,hahn2015sketch,bessmeltsev2016gesture3d,brodt2022sketch2pose,thorne2004motion,yoo2014sketching,igarashi1998path,guay2015space} either target static posing or are limited by database coverage. \textsc{Sketch2Anim} \cite{s2a} introduced multi-conditional diffusion with 2D or 3D keyposes and trajectories for a single human. We extend this paradigm to multi-entity HOH generation with explicit interaction constraints.

\noindent\textbf{Diffusion Control, Flows, Energies, and CTMCs.}
Diffusion backbones remain dominant for motion generation or editing \cite{tevet2023human,chen2023executing,jiang2023motiongpt}, but precise control often relies on posterior guidance \cite{chung2022diffusion} or ControlNet-like modules \cite{xie2023omnicontrol,zhong2025smoodi}, which increase sampling cost and can underperform on hard constraints or multi-track coupling. Rectified or conditional flows \cite{liu2022flow,lipman2022flow} offer direct transport under strong conditions, and recent work shows fast flow-based sampling with strong conditional alignment \cite{ben2024d,esser2024scaling,motionlab}. Energy-based models encode structured priors \cite{du2019implicit,energy_work}, and CTMCs provide a natural mechanism for discrete interaction states \cite{CTMC}. Our method combines these ingredients for sketch-conditioned \emph{multi-entity} HOH generation. Additional preliminaries, including notation, rectified flows, CTMCs, and energies, appear in the Supplementary (Supp.~Sec.~S1-2).

\vspace{-2mm}

\section{Method}
\label{sec:method}
\vspace{-1mm}
We synthesize an $N$-frame, temporally coherent 3D motion sequence for a scene with $H$ humans and $O$ objects, driven by a storyboard of $K$ keyframes and optional text. We use four time variables: decoded motion frames use $n\in\{1,\ldots,N\}$, storyboard keyframe locations use $k\in\mathcal{T}_{\text{key}}\subset[0,N]$ with $k=0$ and $k=N$ denoting the first and last motion frames, the rectified-flow student uses continuous time $t\in[0,1]$, and the diffusion teacher uses diffusion time $\tau\in[0,1]$. From the storyboard we extract 2D controls: for each human $h$ we obtain keyposes at annotated storyboard locations $\mathcal{T}_{\text{key}}$ as $\mathbf{K}^{(h)}_{2\text{D}}[k]\in\mathbb{R}^{J\times 2}$, and, when provided, per-joint 2D polylines describing target image-plane trajectories $\mathbf{T}^{(h,j)}_{2\text{D}}\in\mathbb{R}^{L_{h,j}\times 2}$. For objects we use binary silhouettes or coarse masks $\mathbf{S}^{(o)}_{2\text{D}}[k]\in\{0,1\}^{H_s\times W_s}$ to indicate placement and extent. An optional free-form text prompt is denoted by $\mathbf{a}$. We group these conditioning signals as $\mathcal{C}=(\mathbf{K}_{2\text{D}},\mathbf{T}_{2\text{D}},\mathbf{S}_{2\text{D}},\mathbf{a})$ and maintain a per-frame joint-conditioning indicator $\boldsymbol{\chi}\in\{0,1\}^{N\times H\times J}$, where $\chi_{n,h,j}=1$ indicates that joint $j$ of human $h$ is constrained at frame $n$. When a storyboard location $k\in\mathcal{T}_{\text{key}}$ indexes a decoded-frame quantity, we map it to the nearest frame by $\nu(k)=1+\operatorname{round}\!\left({k(N-1)}/{N}\right)\in\{1,\dots,N\}
$. Object control is conveyed via $\mathbf{S}_{2\text{D}}$ together with designated anchor and contact assignments in the interaction energies.

We write the full scene state at frame $n$ as $\mathbf{M}_n=\mathbf{M}^{\text{hum}}_n\cup\mathbf{M}^{\text{obj}}_n$. For each human $h\in\{1,\ldots,H\}$ we track a kinematic skeleton with $J{=}22$ body joints from a SMPL-X-derived subset \cite{pavlakos2019expressive}. Joint positions are $\mathbf{P}_n^{(h)}\in\mathbb{R}^{J\times 3}$ in a world frame with ground plane $z{=}0$, and joint orientations use the continuous 6D rotation representation \cite{zhou2019continuity} as $\mathbf{R}_n^{(h)}\in\mathbb{R}^{J\times 6}$. For each object $o\in\{1,\ldots,O\}$ we store $K_o$ designated world-space anchors as $\mathbf{Y}_n^{(o)}\in\mathbb{R}^{K_o\times 3}$ and a rigid body pose $\mathbf{O}_n^{(o)}\in\mathbb{R}^{7}$ implemented as 3D translation plus unit quaternion. The flattened scene state has dimension
\vspace{-7mm}
{\small\begin{equation}
D_{\text{full}} = H\cdot J\cdot(3+6) + O\cdot(3+4) + \sum_{o=1}^{O}K_o\cdot 3,
\vspace{-4mm}
\end{equation}}
so $\mathbf{M}_n\in\mathbb{R}^{D_{\text{full}}}$. We denote the stacked sequence by $\mathbf{M}_{1:N}:=[\mathbf{M}_1,\ldots,\mathbf{M}_N]\in\mathbb{R}^{N\times D_{\text{full}}}$ and seek to generate this dense motion given the sparse storyboard controls. Projection to and from the sketch plane follows \cite{s2a}.

Following \cite{collage}, we operate in a hierarchical VQ-VAE \emph{entity} latent.
At level $\ell\in\{1,\ldots,L\}$ the frozen encoder outputs continuous latents
$\mathbf{z}^{(\ell)}\in\mathbb{R}^{T_{\text{lat}}\times V\times d}$ and nearest-code indices
$\mathbf{c}^{(\ell)}\in\{1,\ldots,K_\ell\}^{T_{\text{lat}}\times V}$, where
$V=H+O$ is the number of entity tokens and $T_{\text{lat}}\le N$ is the latent temporal length after sub-sampling.
We combine the level-wise latents as in \cite{collage} and use the resulting continuous latent
$\mathbf{z}=\sum_{\ell=1}^{L}\mathbf{z}^{(\ell)}
\in\mathbb{R}^{T_{\text{lat}}\times V\times d}$.

The frozen decoder $\mathcal{D}:\mathbb{R}^{T_{\text{lat}}\times V\times d}\!\rightarrow\!\mathbb{R}^{N\times D_{\text{full}}}$
upsamples temporally and decodes per-entity motion. Because guidance is defined in motion space but sampling occurs in latent space,
we learn a low-rank block-Toeplitz Jacobian surrogate
$\mathbf{B}_\rho \approx (\partial \mathcal{D}/\partial \mathbf{z})^\top \in \mathbb{R}^{(T_{\text{lat}}\cdot V\cdot d)\times (N\cdot D_{\text{full}})}$
to precondition gradients stably.

To bridge sketches and 3D motion, similar to Sketch2Anim~\cite{s2a,radford2021learning,tevet2022motionclip}, we train paired 2D or 3D encoders for keyposes, trajectories, and objects, $\{\mathcal{E}^{2\text{D}}_y,\mathcal{E}^{3\text{D}}_y\}_{y\in\{\mathrm{kf},\mathrm{tr},o\}}$, each mapping their inputs to a shared embedding space $\mathbb{R}^{d_e}$ with $d_e{=}256$. The alignment loss
\vspace{-2mm}
{\small \begin{equation}
\mathcal{L}_{\text{align}}=\sum_{y\in\{\mathrm{kf},\mathrm{tr},o\}}\|\mathbf{s}^{3\text{D}}_y-\mathbf{s}^{2\text{D}}_y\|_2^2+\lambda_c\mathcal{L}_{\text{InfoNCE}}(y)
\label{eq:align_loss}
\vspace{-2mm}
\end{equation}}
encourages modality-invariant embeddings by combining an $L_2$ term and a mini-batch contrastive objective \cite{oord2018representation}. At inference we retain only the 2D branches, feeding the generator via attention and feature adapters described below.

\begin{figure*}[t]
\centering
\includegraphics[width=\linewidth]{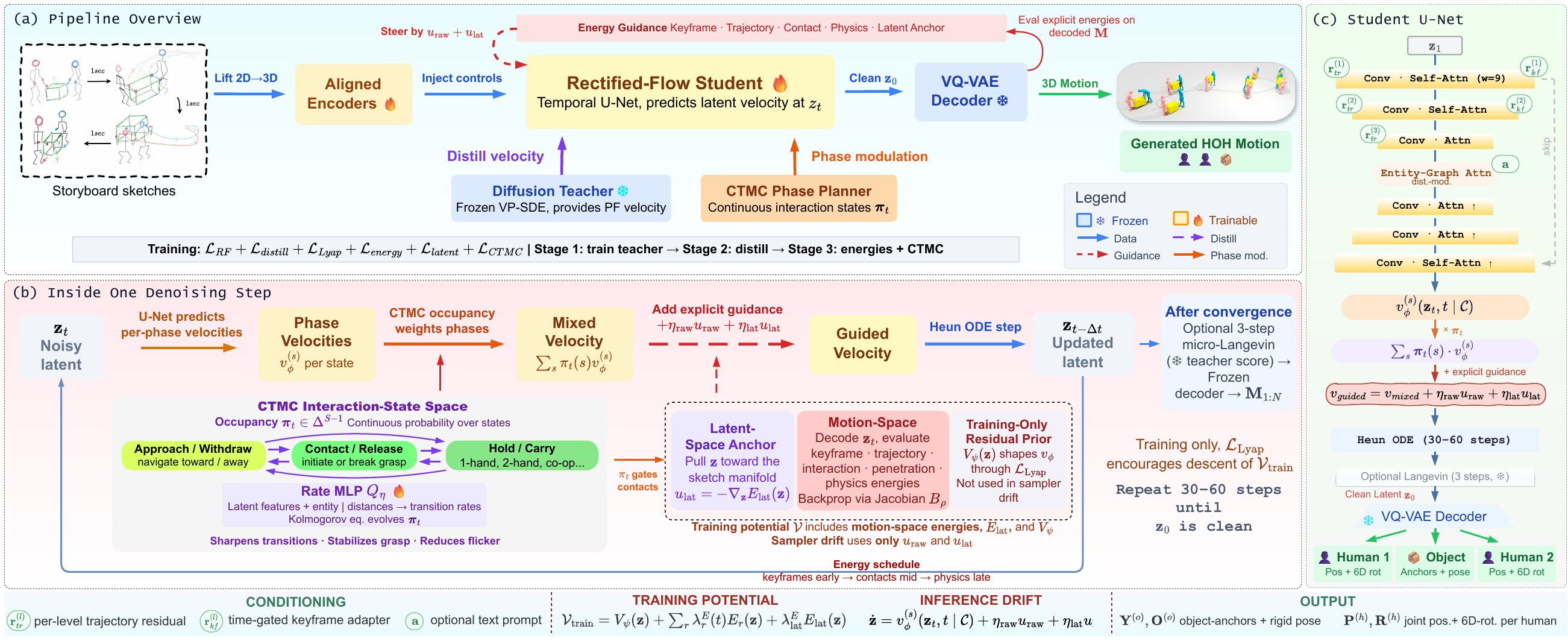}
\vspace{-8mm}
\caption{\textbf{Overview of \ourmodel.} \textbf{(a)} Storyboard sketches and optional text are encoded as controls for latent motion generation. A diffusion teacher guides a rectified-flow student, and a frozen decoder maps the final latent to 3D motion. \textbf{(b)} At each update step, the student predicts a motion change, the phase planner selects the current interaction stage, and guidance keeps the motion close to the sketched poses, paths, and contacts. \textbf{(c)} The student is a temporal U-Net with entity attention for coordinated multi-entity motion.}
\vspace{-6mm}
\label{fig:our_model_main}
\end{figure*}

\vspace{-1mm}
\subsection{Generative Field: Diffusion Teacher and Rectified-Flow Student}
\label{subsec:gen-field}
\vspace{-0.5mm}
Both the diffusion teacher and the rectified-flow student share the same backbone: a 4-level temporal U-Net \cite{collage,ho2020denoising} acting on $\mathbf{z}\!\in\!\mathbb{R}^{T_{\text{lat}}\times V\times d}$ with temporal convolutions \cite{bai2018empirical}, local temporal self-attention, and an entity-graph attention block whose logits are modulated by metric distances between entity tokens. Let $\mathbf{Q},\mathbf{K},\mathbf{V}\!\in\!\mathbb{R}^{T_{\text{lat}}\times V\times d_h}$ be head-wise projections and let $\mathbf{D}_q\!\in\!\mathbb{R}^{V\times V}$ be the pairwise distance matrix at latent step $q\in\{1,\ldots,T_{\text{lat}}\}$, with zero diagonal, computed from cached low-resolution decodes. We apply attention at each latent step and stack:
\vspace{-2mm}
{\small \begin{equation}
\text{Attn}_q \!=\! \mathrm{Softmax}\!\Big(\tfrac{\mathbf{Q}_q \mathbf{K}_q^\top}{\sqrt{d_h}} - \lambda \mathbf{D}_q\Big)\mathbf{V}_q
\label{eq:dist_attn}
\vspace{-1mm}
\end{equation}}
This biases information flow toward spatially proximal agents and objects at each latent step \cite{collage}.

\vspace{-\baselineskip}
\paragraph{Teacher probability flow.}
\textbf{We first pre-train the diffusion \emph{teacher} to convergence and then \emph{freeze} its parameters $\theta$. All subsequent teacher-based supervision uses the teacher's fixed probability-flow velocity.}
The diffusion teacher follows a variance-preserving (VP) forward process at diffusion time $\tau\in[0,1]$,
$\mathbf{z}_\tau=\sqrt{\bar{\alpha}_\tau}\mathbf{z}_0+\sqrt{1-\bar{\alpha}_\tau}\boldsymbol{\epsilon}$ with $\boldsymbol{\epsilon}\sim\mathcal{N}(0,\mathbf{I})$, and predicts noise $\hat{\boldsymbol{\epsilon}}_\theta(\mathbf{z}_\tau,\tau\mid\mathcal{C})$. Its probability-flow ODE velocity is
\vspace{-3mm}
{\small \begin{equation}
v_\theta^{\text{PF}}(\mathbf{z}_\tau,\tau\mid\mathcal{C})
=
-\frac{1}{2}\beta(\tau)\left(
\mathbf{z}_\tau-
\frac{\hat{\boldsymbol{\epsilon}}_\theta(\mathbf{z}_\tau,\tau\mid\mathcal{C})}{\sqrt{1-\bar{\alpha}_\tau}}
\right),
\vspace{-2mm}
\label{eq:pf_velocity}
\end{equation}}
where $\bar{\alpha}_\tau=\exp(-\int_0^\tau \beta(s)\,ds)$. Because rectified-flow time $t$ and diffusion time $\tau$ trace different probability paths, distillation uses a monotone map $\tau=\mathcal{T}(t)$ (details in Supp.~Sec.~S2) and evaluates the frozen teacher PF field at the current student state $\mathbf{z}_t$. The student minimizes the rectified-flow objective \cite{liu2022flow}
\vspace{-4mm}
\begin{equation}
\mathcal{L}_{\text{RF}}(\phi) = \mathbb{E}_{t,\mathbf{z}_0,\mathbf{z}_1} \left\| \mathbf{v}_\phi(\mathbf{z}_t, t \mid \mathcal{C}) - (\mathbf{z}_1 - \mathbf{z}_0) \right\|_2^2 ,
\vspace{-1mm}
\label{eq:rf_loss}
\end{equation}
where $\mathbf{z}_t = (1-t)\mathbf{z}_0 + t\mathbf{z}_1$, together with explicit PF distillation
\vspace{-3.5mm}
{\small \begin{equation}
\mathcal{L}_{\text{distill}}(\phi)=\mathbb{E}_{t,\mathbf{z}_t}\big\|v_\phi(\mathbf{z}_t,t\mid\mathcal{C})-v_\theta^{\text{PF}}(\mathbf{z}_t,\mathcal{T}(t)\mid\mathcal{C})\big\|_2^2,
\vspace{-1mm}
\label{eq:distill_loss}
\end{equation}}

We use a \emph{lift-then-fuse} scheme tailored to storyboard control. The aligned encoders from Eq.~\eqref{eq:align_loss} map 2D human keyposes and joint trajectories to 3D proxy controls $(\hat{\mathbf{K}}_{3\text{D}},\hat{\mathbf{T}}_{3\text{D}})$. Object masks $\mathbf{S}_{2\mathrm{D}}$ are embedded by the object branch and used for object association and the latent anchor path. Following \cite{s2a,xie2023omnicontrol}, we condition the latent U-Net through two routes: \textbf{(i)} a \emph{trajectory path} that injects per-level residuals $\mathbf{r}^{(\ell)}_{\mathrm{tr}}(q)$ from $\hat{\mathbf{T}}_{3\text{D}}$ and adds a temporal attention bias $\mathbf{b}^{(\ell)}_{\mathrm{tr}}$ computed from along-path phase $\boldsymbol{\varphi}_q=[s_q,\dot{s}_q,\kappa_q]$, and \textbf{(ii)} a \emph{time-gated keyframe adapter} that applies localized residuals $\mathbf{r}^{(\ell)}_{\mathrm{kf}}(q)=\sum_{k^\star\in\mathcal{T}_{\mathrm{key}}}\delta_{\sigma}((q-1)/(T_{\text{lat}}-1)-k^\star/N)\,\mathcal{F}^{3\mathrm{D}}_{\mathrm{kf}}\!\big(\hat{\mathbf{K}}_{3\text{D}}[k^\star]\big)$ near key locations $\mathcal{T}_{\mathrm{key}}$. Text $\mathbf{a}$ is encoded once and applied by a single bottleneck cross-attention \cite{collage}. Optionally, as in \cite{s2a}, a weak \emph{trajectory-only} 2D projection guidance can be used at inference to nudge projected end-effector paths toward $\mathbf{T}_{2\text{D}}$ when sketches are reliable. Reported quantitative results do not use this optional correction. We use classifier-free guidance with $10\%$ conditional dropout during training and guidance weight $\omega\in[1.4,1.8]$ at inference.
\vspace{-2mm}
\subsection{Dual-Space Conditioning}
\label{subsec:dual}
\vspace{-1mm}
Sketch constraints are sparse and local. Raw-space guidance provides geometric precision but can drift off-manifold, whereas latent-space guidance preserves coherence but is less precise. We therefore combine both.

Our approach operates through two coupled pathways. First, we define differentiable energies on decoded motions $\Pi(\mathbf{z})=\mathcal{D}(\mathbf{z})$ (detailed in Sec.~\ref{subsec:guidance}) and back-propagate their gradients to latent space using the learned low-rank, block-Toeplitz Jacobian surrogate $\mathbf{B}_\rho \approx \big(\partial \mathcal{D}/\partial \mathbf{z}\big)^\top$. This yields a raw-space guidance vector:
\vspace{-3mm}
{\small \begin{equation}
u_{\text{raw}}(\mathbf{z}) = - \mathbf{B}_\rho\,\nabla_{\mathbf{M}_{1:N}}\!\left(\sum_r \lambda_r E_r(\Pi(\mathbf{z}))\right) \in  \mathbb{R}^{T_{\text{lat}}\times V\times d}
\vspace{-2mm}
\end{equation}}
For each condition type $y\in\{\mathrm{kf},\mathrm{tr},o\}$, we form training targets from a condition-specific constrained subset. For $y\in\{\mathrm{kf},\mathrm{tr}\}$ we use the human joints and frames selected by $\boldsymbol{\chi}$. For $y=o$ we use the object channels involved in the storyboarded masks, anchors, and rigid poses. We denote the resulting embeddings by $\mathbf{s}^{3\mathrm{D}}_y=\mathcal{E}^{3\mathrm{D}}_y(\hat{\mathbf{M}}^{(y)}_{1:N})$. A lightweight projection head $f_\gamma:\mathbb{R}^{T_{\text{lat}}\times V\times d}\rightarrow\mathbb{R}^{d_e}$ maps the current latent to this embedding space, and during training we optimize
$\mathcal{L}_{\text{lat}}=\sum_y\|f_\gamma(\mathbf{z})-\mathrm{sg}[\mathbf{s}^{3\mathrm{D}}_y]\|_2^2+\lambda_{\text{NCE}}\mathcal{L}_{\text{InfoNCE}}$, where $\mathrm{sg}[\cdot]$ denotes stop-gradient and $\mathcal{L}_{\text{InfoNCE}}$ is evaluated per condition type and summed over $y$. At inference, we use the corresponding frozen 2D storyboard embeddings $\mathbf{s}^{2\mathrm{D}}_y=\mathcal{E}^{2\mathrm{D}}_y(\mathcal{C})$ to define a latent energy
$E_{\text{lat}}(\mathbf{z})=\sum_y\|f_\gamma(\mathbf{z})-\mathbf{s}^{2\mathrm{D}}_y\|_2^2$. This gives raw-space energies for precise contacts and trajectories, while latent anchors suppress off-manifold solutions by maintaining proximity to the sketch-conditioned prior. During training and sampling, we inject $u_{\text{lat}}(\mathbf{z})=-\nabla_{\mathbf{z}}E_{\text{lat}}(\mathbf{z})$ as an auxiliary conservative field into the student drift, scaled by $\eta_{\text{lat}}$, while raw-space energies are routed through $\mathbf{B}_\rho$.
\vspace{-2mm}

\subsection{Energy Guidance and Objective}
\label{subsec:guidance}
\vspace{-1mm}
While latent anchors maintain global coherence, precise satisfaction of sparse sketch constraints requires explicit energy formulations that can navigate the coupled landscape of multi-entity interactions. We therefore define task-specific energies on decoded motion together with a Lyapunov-like potential that cooperates with them.

Let $\Pi(\mathbf{z}){=}\mathcal{D}(\mathbf{z})$ be the decoded sequence and $\mathbf{p}^{(h)}_{n,j}\!\in\!\mathbb{R}^3$ the world-space joint of human $h$ at frame $n$ (projected when needed as $\tilde{\mathbf{p}}^{(h)}_{n,j}{=}\Pi_{\text{cam}}(\mathbf{p}^{(h)}_{n,j})$). From storyboard inputs we obtain lightweight 3D proxies $(\hat{\mathbf{K}}_{3\mathrm{D}},\hat{\mathbf{T}}_{3\mathrm{D}})$ via the aligned 2D$\leftrightarrow$3D encoders from Eq.~\eqref{eq:align_loss}. We combine 3D-proxy terms with time-gated 2D terms. To keep the 2D supervision local and robust, we use small gates in $[0,1]$: a keyframe time gate $g_{\text{time}}(n)$ that localizes 2D keypose penalties around $\mathcal{T}_{\text{key}}$, a keyframe-quality score $g_{\text{qual,key}}(k,h,j)$ that softly down-weights noisy or lift-inconsistent keypoints, and for trajectories a path gate $g_{\text{path}}(n,h,j)$ together with a trajectory-quality score $g_{\text{qual,tr}}(n,h,j)$. We parametrize the effective 2D weights as
{\small
\vspace{-6mm}
\begin{align*}
\lambda^{2\mathrm{D}}_{\text{key}}(k,h,j) &= \bar{\lambda}^{2\mathrm{D}}_{\text{key}}\,g_{\text{time}}(\nu(k))\,g_{\text{qual,key}}(k,h,j), \quad \text{and} \\
\lambda^{2\mathrm{D}}_{\mathrm{tr}}(n,h,j) &= \bar{\lambda}^{2\mathrm{D}}_{\mathrm{tr}}\,g_{\text{path}}(n,h,j)\,g_{\text{qual,tr}}(n,h,j). \\[-7mm] 
\end{align*}
}
All 2D and 3D distances are normalized by per-sequence reference scales, and each energy is normalized by the number of active constraints. Exact gates, normalizations, and the soft polyline distance are given in Supp.~Sec.~S2.\\
\indent For keyframe alignment, $E_{\text{key}}(\mathbf{z})$ is calculated as $E^{3\mathrm{D}}_{\text{key}}{+}E^{2\mathrm{D}}_{\text{key}}$, similarly for trajectories, $E_{\mathrm{tr}}(\mathbf{z})$ is calculated as $E^{3\mathrm{D}}_{\mathrm{tr}}{+}E^{2\mathrm{D}}_{\mathrm{tr}}$, with,
{\small
\vspace{-2mm}
\begin{align*}
E^{3\mathrm{D}}_{\text{key}} &= \lambda^{3\mathrm{D}}_{\text{key}}\!\sum_{k\in\mathcal{T}_{\text{key}}}\!\sum_{h,j}\! \chi_{\nu(k),h,j}\|\mathbf{p}^{(h)}_{\nu(k),j}{-}\hat{\mathbf{K}}^{(h)}_{3\mathrm{D}}[k,j]\|_2^2, \\[-0.7ex]
E^{2\mathrm{D}}_{\text{key}} &= \sum_{k\in\mathcal{T}_{\text{key}}}\!\sum_{h,j}\! \lambda^{2\mathrm{D}}_{\text{key}}(k,h,j)\,\chi_{\nu(k),h,j}\|\tilde{\mathbf{p}}^{(h)}_{\nu(k),j}{-}\mathbf{K}^{(h)}_{2\mathrm{D}}[k,j]\|_2^2. \\[-0.7ex]
E^{3\mathrm{D}}_{\mathrm{tr}} &= \lambda^{3\mathrm{D}}_{\mathrm{tr}}\!\sum_{n,h,j}\! \chi_{n,h,j}\,\mathrm{dist}(\mathbf{p}^{(h)}_{n,j},\hat{\mathbf{T}}^{(h,j)}_{3\mathrm{D}})^2, \\[-0.7ex]
E^{2\mathrm{D}}_{\mathrm{tr}} &= \sum_{n,h,j}\! \lambda^{2\mathrm{D}}_{\mathrm{tr}}(n,h,j)\,\chi_{n,h,j}\,\mathrm{dist}(\tilde{\mathbf{p}}^{(h)}_{n,j},\mathbf{T}^{(h,j)}_{2\mathrm{D}})^2. \\[-6ex]
\end{align*}}

where $\mathrm{dist}(\cdot,\cdot)$ is a soft closest-point distance.

Interaction energies model contact and spacing. For any designated contact pair $(u,v)$ (e.g., a hand joint and an object anchor), we penalize the deviation of the signed distance from a target margin, $E_{\text{int}}=\sum_{n,(u,v)} \psi_\delta(d(\mathbf{x}_{n,u},\mathbf{x}_{n,v})-m)$, with Huber penalty $\psi_\delta$ and distance $d(\cdot,\cdot)$ in world space or along object surface SDFs. Additional spacing terms discourage premature crowding before contact. Finally, physics energies include foot-skating suppression via near-zero horizontal foot velocity during stance phases detected from normalized foot height and vertical velocity, ground-plane constraints, and temporal second-difference smoothness on joint positions and rotations. We schedule the contribution of each term by time-dependent weights $\lambda_r(t)$, prioritizing keyframe hits early, tightening interaction margins mid-trajectory, and enforcing physics late, which reduces oscillatory behavior when all objectives compete uniformly.\\
In addition to these raw-space energies, we reuse the latent anchor from Sec.~\ref{subsec:dual} as an explicit latent-space energy:
$E_{\text{lat}}(\mathbf{z}) = \sum_{y\in\{\mathrm{kf},\mathrm{tr},o\}}\big\|f_\gamma(\mathbf{z})-\mathbf{s}^{2\mathrm{D}}_y\big\|_2^2$,
where $f_\gamma$ maps the current latent $\mathbf{z}$ into the shared embedding space and $\mathbf{s}^{2\mathrm{D}}_y$ are frozen storyboard embeddings from the 2D encoders corresponding to keyposes, trajectories, and objects. This term encourages the latent to stay close to the sketch-conditioned manifold while the raw-space energies refine geometry and contacts. During training, its supervised counterpart $\mathcal{L}_{\text{lat}}$ from Sec.~\ref{subsec:dual} is optimized jointly with the other losses. At inference, $E_{\text{lat}}$ is treated as another energy term inside the total potential.

Beyond these hand-designed energies, we learn a potential $V_\psi(\mathbf{z})$ that captures residual preferences of the motion manifold not expressed by explicit terms. The potential is trained by energy-equilibrium matching \cite{energy_work,du2019implicit,song2021train} with two phases: an optimal-transport warm-up in which we learn to assign lower potential to teacher samples than to latent noise, and a contrastive divergence refinement in which short Langevin chains on $\mathbf{z}$ define negative examples. The student is then trained to produce velocities that are descending directions of the total potential
$\mathcal{V}(\mathbf{z},t)=V_\psi(\mathbf{z})+\sum_r\lambda_r(t)E_r(\mathbf{z})+\lambda_{\text{lat}}E_{\text{lat}}(\mathbf{z})$
by minimizing,
\vspace{-3mm}
\begin{equation}
\label{eq:lyap_loss}
\resizebox{0.9\linewidth}{!}{$
\mathcal{L}_{\text{Lyap}}(\phi,\psi)
=\mathbb{E}_t\left[
\Big(
\max\{0,\nabla_{\mathbf{z}}\mathcal{V}(\mathbf{z},t)\cdot v_\phi(\mathbf{z},t)
+\kappa\|\nabla_{\mathbf{z}}\mathcal{V}(\mathbf{z},t)\|_2^2\}
\Big)^2
\right]
$}
\vspace{-2mm}
\end{equation}
with small $\kappa>0$ as a margin. Gradients of raw-space energies $\nabla_{\mathbf{z}}E_r$ are routed through the surrogate $\mathbf{B}_\rho$, whereas $\nabla_{\mathbf{z}}E_{\text{lat}}$ is computed directly. We use $-\nabla_{\mathbf{z}}\mathcal{V}$ as an additional conservative guidance vector during training, while inference uses the explicit raw-space and latent-space guidance terms described in the supplement. This yields a stable approach to constraint satisfaction while keeping the flow close to the sketch-conditioned latent manifold.
\vspace{-2mm}

\begin{figure*}[t]
\centering
\includegraphics[width=0.99\linewidth]{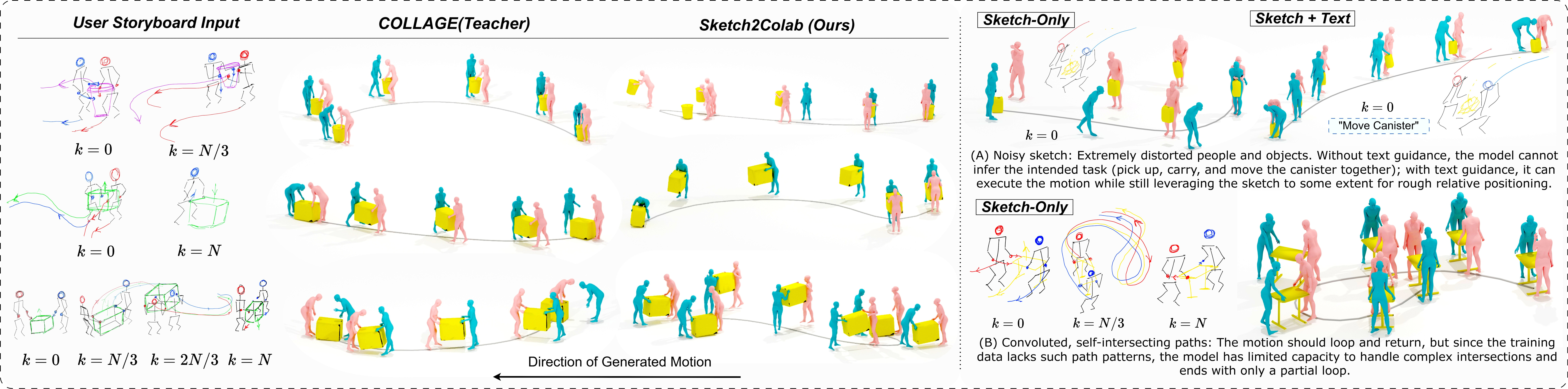}
\vspace{-3mm}
\caption{\textbf{Sketch$\rightarrow$interaction motion.} \textbf{Left panel:} Comparison of \ourmodel and the sketch-conditioned COLLAGE Teacher~\cite{collage}. Given storyboard keyframes and joint trajectories, \ourmodel follows the sketches more faithfully, matches interaction timing, and adheres more closely to keyframes and trajectories. The COLLAGE Teacher often misses storyboard constraints, including the handoff and the continued one-person carry in the first two examples, and the higher lift during motion in the third. \textbf{Right panel:} \textcolor{blue}{(A--B)} Hard cases with heavy sketch noise (\(\approx 60\%\), Table~\ref{tab:main_one_table1}(b)), self-intersecting paths, and sparse constraints that can cause drift, floating, or collisions.}
\label{fig:qual-main}
\vspace{-6mm}
\end{figure*}

\subsection{Discrete Phase Scheduling via CTMC}
\label{subsec:ctmc}
\vspace{-1mm}
Everyday interactions involve switching between a small number of \emph{composite} interaction states whose meaning depends on contact status, relative motion, and active anchor assignments. We therefore couple the continuous flow to a lightweight CTMC over states $s\in\{1,\ldots,S\}$ with occupancy $\boldsymbol{\pi}_t\in\Delta^{S-1}$. In all main experiments we use $S{=}3$ composite states
$\{\text{approach or withdraw},\text{contact or release},\text{hold or carry}\}$,
and we ablate both coarser and finer state granularities in Supp.~Sec.~S5. The first state captures relative navigation before or after stable contact. The second captures transient contact-change events, including grasp onset, release, and handoff transfer. The third captures sustained coupled manipulation, including one-hand carry, two-hand carry, and other cooperative hold regimes. Transition rates are given by a locally equivariant MLP $Q_\eta(\mathbf{h}_t)\in\mathbb{R}^{S\times S}$ with nonnegative off-diagonals and row sums zero, where $\mathbf{h}_t$ are per-time features extracted from the latent, pooled over relevant entity tokens, together with distances to designated anchors. The Kolmogorov forward equation \cite{CTMC}
\vspace{-2.5mm}
{\small\begin{equation}
\frac{d\boldsymbol{\pi}_t}{dt}=\boldsymbol{\pi}_t Q_\eta(\mathbf{h}_t)
\vspace{-2mm}
\end{equation}}
is fit with the core residual loss
\vspace{-3mm}
{\small\begin{equation}
\mathcal{L}_{\text{CTMC}}(\eta)=\mathbb{E}_{t}\left\|\frac{d\boldsymbol{\pi}_t}{dt}-\boldsymbol{\pi}_t Q_\eta(\mathbf{h}_t)\right\|_2^2+\beta\mathrm{Var}[A_t(Q_\eta)],
\label{eq:ctmc_loss}
\vspace{-2mm}
\end{equation}}
where $A_t$ denotes an entropy-like regularizer on the rate spectrum. The full implementation additionally uses weak pseudo-state supervision and a dwell-time regularizer. These details are deferred to Supp.~Sec.~S2.

The CTMC modulates the student in two ways. First, it mixes phase-specific sub-fields,
$\,v_\phi(\mathbf{z},t)=\sum_{s=1}^S \pi_t(s)\,v_\phi^{(s)}(\mathbf{z},t)\,$.
Second, it weights the contact terms in $\mathcal{V}$ by state occupancy, so their influence follows the inferred interaction phase. This is illustrated by the first storyboard in Fig.~\ref{fig:qual-main}. There, \ourmodel first carries the object with two people, then one person peels away while the other continues with the object to the final placement. The shared object and carrier trajectory, together with the diverging trajectory of the second person, cues a mid-sequence support change. The occupancy therefore moves into \textit{hold or carry} during the initial joint transport, returns briefly to \textit{contact or release} when one carrier disengages, and then shifts back to \textit{hold or carry} for the remaining single-person carry before the final release. This sharpens contact changes and stabilizes the carry for our default $S{=}3$ model.

With the teacher frozen, the student (and auxiliary modules $\psi,\eta$) are trained with the combined objective,
\vspace{-3mm}
{\small \begin{equation}
\begin{aligned}
\mathcal{L} =& \mathcal{L}_{\text{RF}}
+\lambda_{\text{dist}}\mathcal{L}_{\text{distill}}
+\lambda_{\text{Lyap}}\mathcal{L}_{\text{Lyap}}
+\sum_r \lambda_r\,\mathcal{L}_{E_r} \\[-4pt]
& +\lambda_{\text{lat}}\mathcal{L}_{\text{lat}}
+\lambda_{\text{CTMC}}\mathcal{L}_{\text{CTMC}}
+\lambda_{\text{cons}}\mathcal{L}_{\text{consist}},
\label{eq:total_loss}
\end{aligned}
\vspace{-2mm}
\end{equation}}
where $\mathcal{L}_{E_r}$ are supervised surrogates for energy terms with ground truth (e.g., contact labels), and $\mathcal{L}_{\text{consist}}$ enforces segment-overlap consistency for long clips per \cite{collage}.
\vspace{1mm}

\vspace{-3mm}
\section{Experimental Setup}
\label{sec:exp-setup}
\vspace{-2mm}

We evaluate \ourmodel\ on multi-entity HOH scenarios with three questions in mind: (i) can the model faithfully follow storyboard keyposes, joint trajectories, and object placements specified by masks and anchors, (ii) does it improve interaction quality (contacts, collisions, spacing) over sketch-driven baselines, and (iii) how does the rectified-flow + CTMC design compare to diffusion-only alternatives in terms of alignment and sampling cost. Below we summarize the datasets, storyboard synthesis, training or inference setup, and evaluation metrics. Additional implementation details are provided in Supp.~Sec.~S4.

\definecolor{hdrR}{HTML}{8E244D}
\definecolor{hdrC}{HTML}{0D47A1}
\definecolor{hdrI}{HTML}{1B5E20}
\definecolor{hdrT}{HTML}{E65100}
\definecolor{bestG}{HTML}{C8E6C9}
\definecolor{rowA}{HTML}{F6F6F6}
\definecolor{oursR}{HTML}{FFF3E0}

\noindent\textbf{Datasets and storyboard synthesis.}
We train and evaluate on \textbf{CORE4D} \cite{zhang2024core4d} and the \textbf{InterHuman} dataset introduced by \cite{liang2023intergen}. For CORE4D, we use the real interaction benchmark subset rather than the full synthetic retargeting branch. This yields $998$ multi-human, object-centric interaction sequences, and we follow the $80/5/15$ train, val, and test split of \cite{collage}. For InterHuman, we use the $6{,}022$-motion benchmark configuration adopted by \cite{collage}. For each 3D motion clip, we synthesize storyboard controls by projecting 3D joints to sketch plane and rasterizing per-joint 2D polylines with perturbations to mimic hand-drawn strokes, building on Sketch2Anim~\cite{s2a}. We also use $682$ hand-drawn sketches over CORE4D sequences. We use the same camera setup as Sketch2Anim~\cite{s2a} and annotate $K$ keyframes per clip at motion-salient times, yielding 2D keyposes $\mathbf{K}_{2\mathrm{D}}$ and 2D joint paths $\mathbf{T}_{2\mathrm{D}}$ as described in Sec.~\ref{sec:method}.

\noindent\textbf{Teacher--student training and inference.}
{\ourmodel} distills a sketch-conditioned diffusion teacher built on the COLLAGE latent backbone and decoder into a rectified-flow student operating in the same latent space (Sec.~\ref{sec:method}). The teacher uses the same storyboard conditioning path described in Sec.~\ref{sec:method}, while retaining the COLLAGE hierarchical VQ-VAE latent representation and decoder. In the main tables, the row labeled \emph{COLLAGE Teacher} denotes this sketch-conditioned diffusion teacher. We first train the teacher with AdamW for $400$k steps (learning rate $2{\times}10^{-4}$, batch size $64$, EMA $0.999$, weight decay $0.01$) using $1000$ diffusion steps \cite{ho2020denoising}, then freeze the teacher and distill its probability-flow field into the student. The student is trained for $500$k steps with learning rate $10^{-4}$ and batch size $64$. In the first phase we minimize $\mathcal{L}_{\text{RF}}+\lambda_{\text{dist}}\mathcal{L}_{\text{distill}}$. In the second phase we activate the Lyapunov and CTMC losses, the supervised energy surrogates, and the latent loss, optimizing the full objective in Eq.~\eqref{eq:total_loss}. Unless otherwise stated we set $\lambda_{\text{dist}}{=}0.5$, $\lambda_{\text{Lyap}}{=}0.5$, $\lambda_{\text{CTMC}}{=}0.1$, $\lambda_{\text{lat}}{=}0.1$, and anneal the energy weights $\lambda_r(t)$ according to the early, mid, and late schedule described in Sec.~\ref{subsec:guidance}. At inference time, we draw $\mathbf{z}_1\sim\mathcal{N}(0,\mathbf{I})$ and, conditioning on the storyboard controls $\mathcal{C}$, integrate the guided ODE for $T_{\text{int}}\!\in\![30,60]$ Heun steps (we use $T_{\text{int}}{=}30$ on CORE4D and $T_{\text{int}}{=}60$ on InterHuman), interleaving CTMC updates every $s{=}3$ steps and using classifier-free guidance weight $\omega\!\in\![1.4,1.8]$ (default $\omega{=}1.6$). After the ODE pass, we optionally apply a short micro-Langevin refinement in latent space: $K_{\text{corr}}{=}3$ Langevin steps with step size $\eta_{\text{corr}}{=}1{\times}10^{-3}$ using the teacher score, which sharpens high-frequency details without altering global motion. The final 3D motion $\mathbf{M}_{1:N}$ is obtained by decoding the final aggregated latent $\mathbf{z}_0$ with the frozen COLLAGE decoder. The student architecture uses the COLLAGE temporal U-Net design \cite{collage}.

\noindent\textbf{Adapted baselines} and \textbf{Metrics.}
For quantitative comparison, \emph{RetrievAdapt} is a retrieval-based baseline that retimes, denoises, and blends retrieved motion to match the storyboard. \emph{Sketch2Anim-INT} is our multi-entity adaptation of single-human \textsc{Sketch2Anim} \cite{s2a}. \emph{COLLAGE Teacher} is our sketch-conditioned adaptation of \textsc{COLLAGE} \cite{collage}, replacing its original text-planning interface with the storyboard conditioning path in Sec.~\ref{sec:method}. These names refer to adapted baselines for our HOH setting, not the original published models. We report realism (FID, Foot-skate), control accuracy (Keypose-2D/3D, Trajectory-2D/3D under the \emph{Average} protocol \cite{s2a}), interaction quality on CORE4D (ObjPos-3D, Anchor-Err, Penetration), and text–motion alignment (MM Dist, R-Precision Top-3 \cite{tevet2022motionclip,collage,s2a}). For InterHuman(human–human only), we report realism, control, penetration, and text–motion alignment but omit object-root and anchor metrics. Full definitions and implementation details are provided in Supp.~Sec.~S4.
\vspace{-2mm}
\section{Results and Analysis}
\label{sec:results}

\vspace{-1mm}
\begin{figure}[t]
\centering
\captionsetup[subfigure]{font=footnotesize, justification=centering}

\begin{subfigure}[t]{0.48\linewidth}
  \centering
  \includegraphics[width=\linewidth]{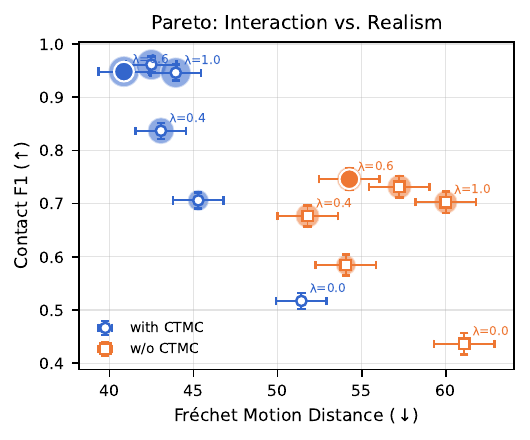}
  \vspace{-6mm}
  \caption{Pareto frontier: F1 vs.~FMD (marker size indicates collision rate).}
  \vspace{-1mm}
  \label{fig:pareto}
\end{subfigure}\hfill
\begin{subfigure}[t]{0.46\linewidth}
  \centering
  \includegraphics[width=\linewidth]{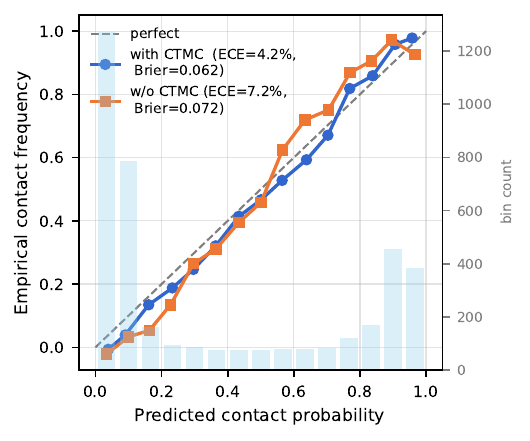}
  \vspace{-6mm}
  \caption{Contact timing calibration (ECE, Brier).}
  \vspace{-2mm}
  \label{fig:reliab}
\end{subfigure}

\begin{subfigure}[t]{0.48\linewidth}
  \centering
  \includegraphics[width=\linewidth]{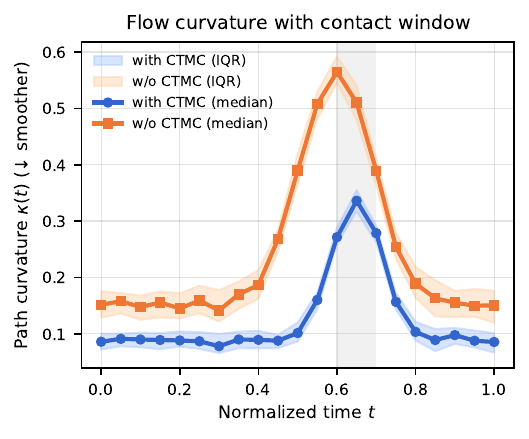}
  \vspace{-6mm}
  \caption{Flow curvature $\kappa(t)$ over time.}
  \label{fig:curv}
\end{subfigure}\hfill
\begin{subfigure}[t]{0.48\linewidth}
  \centering
  \includegraphics[width=\linewidth]{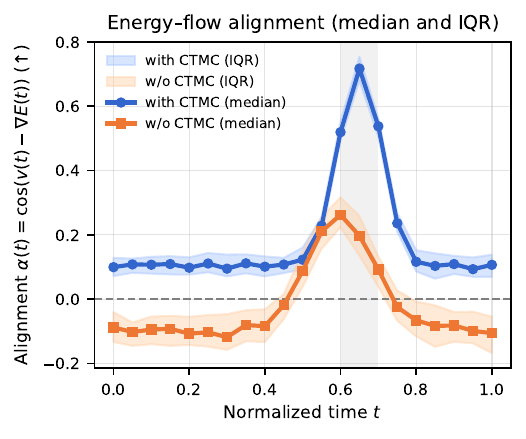}
  \vspace{-6mm}
  \caption{Energy-flow alignment $\alpha(t)$.}
  \label{fig:align}
\end{subfigure}
\vspace{-3mm}
\caption{\textbf{Effect of CTMC scheduling on interaction diagnostics.}
\textbf{(a)} Operating-point sweep over \(\lambda\), showing Contact F1 against Fr\'echet Motion Distance, with marker area proportional to collision rate.
\textbf{(b)} Reliability diagram for contact probability, with Expected Calibration Error and Brier score.
\textbf{(c)} Curvature diagnostic \(\kappa(t)\) over normalized solver time \(t\), shown as median with IQR shading.
\textbf{(d)} Energy-flow cosine alignment \(\alpha(t)\) over normalized solver time \(t\), shown as median with IQR shading.
Error bars in (a) show mean \(\pm\) std over three runs.
The gray band in (c) and (d) marks the annotated contact window.}
\label{fig:diag-2x2}
\vspace{-4mm}
\end{figure}

\begin{table}[t]
\centering
\vspace{1mm}
\caption{\textbf{Storyboard HOH on CORE4D~\cite{zhang2024core4d} under sketch-only conditioning.}
$\downarrow$\,=\,lower is better.
\colorbox{bestG}{\textbf{best}} per panel.
Noise \(60\%\) corresponds to Fig.~\ref{fig:qual-main}A.
\textcolor{hdrR}{\rule{0.7ex}{0.7ex}}\,Realism,
\textcolor{hdrC}{\rule{0.7ex}{0.7ex}}\,Control~\cite{s2a},
\textcolor{hdrI}{\rule{0.7ex}{0.7ex}}\,Interaction.}
\label{tab:main_one_table1}
\vspace{-3mm}
\resizebox{\columnwidth}{!}{%
\setlength{\tabcolsep}{1.5pt}%
\renewcommand{\arraystretch}{0.65}%
\setlength{\aboverulesep}{0.01ex}%
\setlength{\belowrulesep}{0.01ex}%
\begin{tabular}{@{}l|cc|cccc|cc@{}}
\toprule
& \multicolumn{2}{c|}{\cellcolor{hdrR}\textcolor{white}{\scriptsize\textbf{Realism}}}
& \multicolumn{4}{c|}{\cellcolor{hdrC}\textcolor{white}{\scriptsize\textbf{Control Accuracy}}}
& \multicolumn{2}{c}{\cellcolor{hdrI}\textcolor{white}{\scriptsize\textbf{Interaction}}} \\
& {\scriptsize FID}{\tiny$\downarrow$}
& {\scriptsize Ft}{\tiny$\downarrow$}
& {\scriptsize K2D}{\tiny$\downarrow$}
& {\scriptsize K3D}{\tiny$\downarrow$}
& {\scriptsize T2D}{\tiny$\downarrow$}
& {\scriptsize T3D}{\tiny$\downarrow$}
& {\scriptsize Anc}{\tiny$\downarrow$}
& {\scriptsize Pen}{\tiny$\downarrow$} \\
\midrule
\multicolumn{9}{@{}l}{\scriptsize\textit{(a) Method Comparison \& Ablations}} \\
\rowcolor{rowA}
{\scriptsize RetrievAdapt}
  & {\scriptsize .588} & \best{\scriptsize .075}
  & {\scriptsize .074} & {\scriptsize .097} & {\scriptsize .376} & {\scriptsize .525}
  & {\scriptsize .172} & \best{\scriptsize .020} \\
{\scriptsize Sketch2Anim-INT~\cite{s2a}}
  & {\scriptsize .986} & {\scriptsize .148}
  & {\scriptsize .067} & {\scriptsize .084} & {\scriptsize .185} & {\scriptsize .279}
  & {\scriptsize .114} & {\scriptsize .054} \\
\rowcolor{rowA}
{\scriptsize COLLAGE Teacher~\cite{collage}}
  & {\scriptsize .703} & {\scriptsize .143}
  & {\scriptsize .066} & {\scriptsize .082} & {\scriptsize .181} & {\scriptsize .249}
  & {\scriptsize .093} & {\scriptsize .037} \\
\rowcolor{oursR}
{\scriptsize\textbf{\ourmodel}}
  & \best{\scriptsize .487} & {\scriptsize .094}
  & \best{\scriptsize .036} & \best{\scriptsize .050} & \best{\scriptsize .090} & \best{\scriptsize .133}
  & \best{\scriptsize .044} & {\scriptsize .024} \\
\rowcolor{rowA}
{\scriptsize \quad w/o CTMC}
  & {\scriptsize .524} & {\scriptsize .110}
  & {\scriptsize .038} & {\scriptsize .054} & {\scriptsize .100} & {\scriptsize .151}
  & {\scriptsize .059} & {\scriptsize .030} \\
{\scriptsize \quad w/o Energy}
  & {\scriptsize .571} & {\scriptsize .114}
  & {\scriptsize .040} & {\scriptsize .056} & {\scriptsize .114} & {\scriptsize .174}
  & {\scriptsize .090} & {\scriptsize .030} \\
\midrule
\multicolumn{9}{@{}l}{\scriptsize\textit{(b) Sketch-only Noise Robustness}} \\
\rowcolor{oursR}
{\scriptsize\textbf{\ourmodel}}
  & \best{\scriptsize .487} & {\scriptsize .094}
  & \best{\scriptsize .036} & \best{\scriptsize .050} & \best{\scriptsize .090} & \best{\scriptsize .133}
  & \best{\scriptsize .044} & {\scriptsize .024} \\
\rowcolor{rowA}
{\scriptsize \quad Noise 20\%}
  & {\scriptsize .527} & {\scriptsize .115}
  & {\scriptsize .042} & {\scriptsize .059} & {\scriptsize .113} & {\scriptsize .170}
  & {\scriptsize .064} & {\scriptsize .033} \\
{\scriptsize \quad Noise 40\%}
  & {\scriptsize .572} & {\scriptsize .121}
  & {\scriptsize .047} & {\scriptsize .066} & {\scriptsize .133} & {\scriptsize .205}
  & {\scriptsize .087} & {\scriptsize .040} \\
\rowcolor{rowA}
{\scriptsize \quad Noise 60\%}
  & {\scriptsize .698} & {\scriptsize .145}
  & {\scriptsize .060} & {\scriptsize .085} & {\scriptsize .172} & {\scriptsize .275}
  & {\scriptsize .129} & {\scriptsize .050} \\
{\scriptsize \quad Noise 80\%}
  & {\scriptsize .856} & {\scriptsize .166}
  & {\scriptsize .075} & {\scriptsize .110} & {\scriptsize .230} & {\scriptsize .365}
  & {\scriptsize .178} & {\scriptsize .069} \\
\bottomrule
\end{tabular}}
\vspace{-7mm}
\end{table}

\begin{figure*}[t]
\centering
\includegraphics[width=\linewidth]{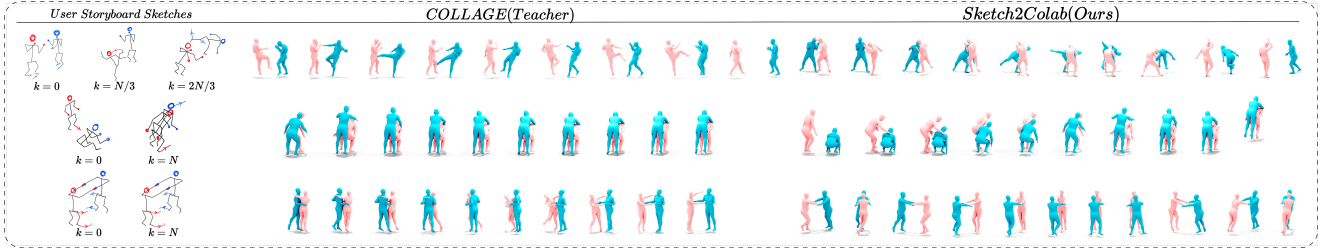}
\vspace{-7mm}
\caption{\textbf{Sketch-only comparisons to the sketch-conditioned COLLAGE Teacher on InterHuman~\cite{liang2023intergen} (HH).} \ourmodel follows the sketched poses, timing, and role changes more faithfully. In the first example it matches the overhead kick and its timing, and in the second it captures the intended crouch swap before one person carries the other. The COLLAGE Teacher produces plausible motion, but it often misses the intended keyframe timing and follows the sketches less precisely.}
\label{fig:hhfail}
\vspace{-5.5mm}
\end{figure*}

\noindent\textbf{Overview.}
Table~\ref{tab:main_one_table} reports the sketch+text setting on \textbf{CORE4D} across realism, control accuracy, interaction quality, and text–motion alignment. Table~\ref{tab:main_one_table1} reports the corresponding sketch-only setting. In the sketch+text setting, \ourmodel establishes the strongest overall operating point for storyboard-driven HOH generation: it improves realism, tightens both keyframe and trajectory adherence in 2D and 3D, materially improves object or anchor interaction while suppressing penetrations, and strengthens text–motion alignment. Relative to COLLAGE Teacher, \ourmodel cuts FID by roughly \textbf{22\%} and foot-skate by about \textbf{24\%}. Key-2D or 3D errors drop by around \textbf{37\%} or \textbf{32\%}, Traj-2D by about \textbf{45\%}, and Traj-3D by about \textbf{40\%}. On interaction quality, object-position and anchor errors decrease by about \textbf{45\%} and \textbf{51\%}, with penetration reduced by about \textbf{21\%}. Text–motion alignment also improves, with MM Dist decreasing by about \textbf{10\%} and RP3 rising from $0.500$ to $0.522$. Table~\ref{tab:main_one_table1} further shows that the sketch-only model already beats all baselines on HOH storyboard control, while optional text gives modest gains for clean sketches and larger gains as sketch noise increases.

\noindent\textbf{Comparison to adapted baselines.}
Against Sketch2Anim-INT, \ourmodel reduces FID by roughly \textbf{45\%}, Traj-3D by about \textbf{48\%}, anchor error by about \textbf{58\%}, and penetration by about \textbf{49\%}, while also improving keyframe control and object alignment. These gains come from operating in COLLAGE’s structured latent space and refining it with dual-space guidance rather than relying on raw-space control alone. RetrievAdapt yields the lowest foot-skate and penetration, consistent with its retrieval-plus-refinement design. Its main limitation is storyboard fidelity: it underperforms the learned baselines on trajectory following and object or anchor alignment, with larger gaps in the sketch-only HOH setting (Table~\ref{tab:main_one_table1}). In contrast, \ourmodel retains near-retrieval smoothness while following storyboard constraints and interaction intent much more faithfully.

\begin{table}[t]
\centering
\caption{\textbf{Quantitative results on CORE4D~\cite{zhang2024core4d} (Average protocol, sketch+text unless noted).}
$\downarrow$\,=\,lower, $\uparrow$\,=\,higher is better.
\colorbox{bestG}{\textbf{best}} per panel.
Baselines and ablations in (a), sketch+text noise robustness in (b).
\textcolor{hdrR}{\rule{0.7ex}{0.7ex}}\,Realism,
\textcolor{hdrC}{\rule{0.7ex}{0.7ex}}\,Control~\cite{s2a},
\textcolor{hdrI}{\rule{0.7ex}{0.7ex}}\,Interaction,
\textcolor{hdrT}{\rule{0.7ex}{0.7ex}}\,Text--Motion.
Cross-protocol and InterHuman~\cite{liang2023intergen} results in Supp.\,S.5.}
\label{tab:main_one_table}
\vspace{-3mm}
\resizebox{\columnwidth}{!}{%
\setlength{\tabcolsep}{1pt}%
\renewcommand{\arraystretch}{0.7}%
\setlength{\aboverulesep}{0.15ex}%
\setlength{\belowrulesep}{0ex}%
\begin{tabular}{@{}l|cc|cccc|ccc|cc@{}}
\toprule
& \multicolumn{2}{c|}{\cellcolor{hdrR}\textcolor{white}{\scriptsize\textbf{Realism}}}
& \multicolumn{4}{c|}{\cellcolor{hdrC}\textcolor{white}{\scriptsize\textbf{Control Accuracy}}}
& \multicolumn{3}{c|}{\cellcolor{hdrI}\textcolor{white}{\scriptsize\textbf{Interaction}}}
& \multicolumn{2}{c}{\cellcolor{hdrT}\textcolor{white}{\scriptsize\textbf{Text--Mot.}}} \\
& {\scriptsize FID}{\tiny$\downarrow$}
& {\scriptsize Ft}{\tiny$\downarrow$}
& {\scriptsize K2D}{\tiny$\downarrow$}
& {\scriptsize K3D}{\tiny$\downarrow$}
& {\scriptsize T2D}{\tiny$\downarrow$}
& {\scriptsize T3D}{\tiny$\downarrow$}
& {\scriptsize Obj}{\tiny$\downarrow$}
& {\scriptsize Anc}{\tiny$\downarrow$}
& {\scriptsize Pen}{\tiny$\downarrow$}
& {\scriptsize MM Dist}{\tiny$\downarrow$}
& {\scriptsize RP3}{\tiny$\uparrow$} \\
\midrule
\multicolumn{12}{@{}l}{\scriptsize\textit{(a) Baselines \& Ablations}} \\
\rowcolor{rowA}
{\scriptsize RetrievAdapt}
  & {\scriptsize .469} & \best{\scriptsize .071}
  & {\scriptsize .054} & {\scriptsize .068} & {\scriptsize .255} & {\scriptsize .356}
  & {\scriptsize .081} & {\scriptsize .121} & \best{\scriptsize .019}
  & {\scriptsize 6.41} & {\scriptsize .463} \\
{\scriptsize Sketch2Anim-INT~\cite{s2a}}
  & {\scriptsize .728} & {\scriptsize .118}
  & {\scriptsize .053} & {\scriptsize .066} & {\scriptsize .145} & {\scriptsize .222}
  & {\scriptsize .063} & {\scriptsize .091} & {\scriptsize .043}
  & {\scriptsize 6.23} & {\scriptsize .477} \\
\rowcolor{rowA}
{\scriptsize COLLAGE Teacher~\cite{collage}}
  & {\scriptsize .511} & {\scriptsize .111}
  & {\scriptsize .051} & {\scriptsize .063} & {\scriptsize .140} & {\scriptsize .194}
  & {\scriptsize .062} & {\scriptsize .077} & {\scriptsize .028}
  & {\scriptsize 6.08} & {\scriptsize .500} \\
\rowcolor{oursR}
{\scriptsize\textbf{\ourmodel}}
  & \best{\scriptsize .399} & {\scriptsize .084}
  & \best{\scriptsize .032} & \best{\scriptsize .043} & \best{\scriptsize .078} & \best{\scriptsize .116}
  & \best{\scriptsize .034} & \best{\scriptsize .038} & {\scriptsize .022}
  & \best{\scriptsize 5.50} & \best{\scriptsize .522} \\
\midrule
\rowcolor{rowA}
{\scriptsize \quad w/o Energy}
  & {\scriptsize .471} & {\scriptsize .098}
  & {\scriptsize .035} & {\scriptsize .048} & {\scriptsize .099} & {\scriptsize .149}
  & {\scriptsize .042} & {\scriptsize .075} & {\scriptsize .026}
  & {\scriptsize 5.59} & {\scriptsize .514} \\
{\scriptsize \quad w/o CTMC}
  & {\scriptsize .432} & {\scriptsize .099}
  & {\scriptsize .034} & {\scriptsize .046} & {\scriptsize .086} & {\scriptsize .132}
  & {\scriptsize .039} & {\scriptsize .050} & {\scriptsize .027}
  & {\scriptsize 5.54} & {\scriptsize .518} \\
\rowcolor{rowA}
{\scriptsize \quad w/o Coll.\ grounding}
  & {\scriptsize .420} & {\scriptsize .088}
  & {\scriptsize .033} & {\scriptsize .044} & {\scriptsize .080} & {\scriptsize .125}
  & {\scriptsize .042} & {\scriptsize .047} & {\scriptsize .025}
  & {\scriptsize 5.53} & {\scriptsize .521} \\
{\scriptsize \quad w/o Temporal bias}
  & {\scriptsize .415} & {\scriptsize .088}
  & {\scriptsize .033} & {\scriptsize .044} & {\scriptsize .082} & {\scriptsize .122}
  & {\scriptsize .036} & {\scriptsize .043} & {\scriptsize .022}
  & {\scriptsize 5.51} & {\scriptsize .520} \\
\rowcolor{rowA}
{\scriptsize \quad Parallel ControlNets}
  & {\scriptsize .440} & {\scriptsize .092}
  & {\scriptsize .034} & {\scriptsize .045} & {\scriptsize .090} & {\scriptsize .128}
  & {\scriptsize .040} & {\scriptsize .048} & {\scriptsize .023}
  & {\scriptsize 5.56} & {\scriptsize .519} \\
{\scriptsize \quad Trajectory-only}
  & {\scriptsize .484} & {\scriptsize .097}
  & {\scriptsize .037} & {\scriptsize .049} & {\scriptsize .104} & {\scriptsize .149}
  & {\scriptsize .044} & {\scriptsize .061} & {\scriptsize .024}
  & {\scriptsize 5.61} & {\scriptsize .514} \\
\midrule
\multicolumn{12}{@{}l}{\scriptsize\textit{(b) Sketch+Text Noise Robustness}} \\
\rowcolor{oursR}
{\scriptsize\textbf{\ourmodel}}
  & \best{\scriptsize .399} & {\scriptsize .084}
  & \best{\scriptsize .032} & \best{\scriptsize .043} & \best{\scriptsize .078} & \best{\scriptsize .116}
  & \best{\scriptsize .034} & \best{\scriptsize .038} & {\scriptsize .022}
  & \best{\scriptsize 5.50} & \best{\scriptsize .522} \\
\rowcolor{rowA}
{\scriptsize \quad Noise 20\%}
  & {\scriptsize .414} & {\scriptsize .090}
  & {\scriptsize .035} & {\scriptsize .047} & {\scriptsize .088} & {\scriptsize .132}
  & {\scriptsize .038} & {\scriptsize .046} & {\scriptsize .023}
  & {\scriptsize 5.55} & {\scriptsize .518} \\
{\scriptsize \quad Noise 40\%}
  & {\scriptsize .451} & {\scriptsize .098}
  & {\scriptsize .038} & {\scriptsize .052} & {\scriptsize .106} & {\scriptsize .163}
  & {\scriptsize .044} & {\scriptsize .064} & {\scriptsize .029}
  & {\scriptsize 5.64} & {\scriptsize .510} \\
\rowcolor{rowA}
{\scriptsize \quad Noise 60\%}
  & {\scriptsize .546} & {\scriptsize .117}
  & {\scriptsize .045} & {\scriptsize .065} & {\scriptsize .137} & {\scriptsize .214}
  & {\scriptsize .055} & {\scriptsize .095} & {\scriptsize .034}
  & {\scriptsize 5.82} & {\scriptsize .497} \\
{\scriptsize \quad Noise 80\%}
  & {\scriptsize .650} & {\scriptsize .127}
  & {\scriptsize .056} & {\scriptsize .081} & {\scriptsize .175} & {\scriptsize .279}
  & {\scriptsize .070} & {\scriptsize .126} & {\scriptsize .050}
  & {\scriptsize 6.05} & {\scriptsize .479} \\
\bottomrule
\end{tabular}}
\vspace{-7mm}
\end{table}

\noindent\textbf{Ablations.}
The ablation block in Table~\ref{tab:main_one_table} isolates the contribution of each component. Removing \emph{energy guidance} produces the largest drop: FID rises by about \textbf{18\%}, Traj-3D by about \textbf{28\%}, object error by about \textbf{24\%}, anchor error nearly \textbf{doubles}, and penetration worsens by about \textbf{18\%}, confirming that the raw-space energies are the primary driver of precise, safe contacts. Disabling \emph{CTMC scheduling} mainly hurts temporal phasing. Foot-skate rises by about \textbf{18\%}, Traj-3D by about \textbf{14\%}, anchor error by about \textbf{32\%}, and penetration by about \textbf{23\%}, consistent with its role in staging approach, contact, and hold transitions. Removing \emph{COLLAGE latent grounding}(w/o Coll. grounding) causes more focused regressions in layout and object interaction, with object and anchor errors increasing by about \textbf{24\%} and \textbf{24\%}. Ablating the \emph{temporal bias} yields smaller but systematic drops concentrated on trajectories and anchor timing. Replacing the unified conditioning path with \emph{parallel ControlNets}\cite{s2a} causes broad moderate degradation, while a \emph{trajectory-only} route most severely erodes combined control and interaction, increasing Traj-2D or 3D by about \textbf{33\%} or \textbf{28\%} and anchor error by about \textbf{61\%}. Additional ablations are provided in Supp.~Sec.~S5.

\noindent\textbf{Qualitative Analysis.}
The qualitative panels in Fig.~\ref{fig:qual-main},~\ref{fig:hhfail} and the supplement mirror the table trends: \ourmodel follows sketched paths tightly, aligns with specified anchors, and generates poses that closely match storyboard controls. The model executes phase transitions at the intended times and maintains crisp object alignment and interaction without accumulating spurious motion. In contrast, adapted baselines drift, over-constrain too early or too weakly, and frequently miss intended handoffs or contact timing. We also observe that retrieval-based generation remains smooth but is much less faithful to detailed sketch constraints. Visualizations are available on papers website.
\vspace{-3mm}
\section{Discussion and Future Work}
\vspace{-2mm}
We introduced \ourmodel, a storyboard-driven rectified-flow framework with CTMC phase scheduling, dual-space guidance, and latent anchors that together yield strong realism, control accuracy, and interaction quality for HOH generation while retaining efficient sampling. Our current implementation focuses primarily on two-human collaborative interactions with a single actively manipulated object at a time, rather than denser multi-human coordination or multi-object task graphs, and it still inherits part of its behavior from a calibrated diffusion teacher. In future work, we aim to support broader object categories given only geometry and physical properties, extend beyond two humans and single-object collaborations via compositional field models, and reduce dependence on a diffusion prior.

{
    \small
    \bibliographystyle{ieeenat_fullname}
    \bibliography{main}
}
\end{document}